\documentclass[letterpaper, 10 pt, conference]{ieeeconf}
\IEEEoverridecommandlockouts   
\usepackage{lipsum} 
\usepackage[T1]{fontenc}
\usepackage{graphicx}
\usepackage{graphics}
\usepackage{multirow}
\usepackage{float}
\usepackage{hyperref}
\usepackage{epstopdf}
\usepackage{enumerate}
\usepackage{multicol,multirow}
\usepackage{booktabs}
\usepackage{cite}
\usepackage{amsmath}
\usepackage{academicons}
\usepackage{amssymb}
\usepackage{amsfonts}
\usepackage{bbm}
\usepackage{balance}
\usepackage{paralist}
\usepackage[binary-units=true]{siunitx}
\graphicspath{{figures/}}
\usepackage[ruled,vlined,linesnumbered]{algorithm2e}
\let\oldnl\nl% Store \nl in \oldnl
\newcommand{\nonl}{\renewcommand{\nl}{\let\nl\oldnl}}

\graphicspath{{figures/}} 

\begin{document}
	\title{\LARGE\bf
	  Exploiting Spherical Projections To Generate Human-Like Wrist Pointing Movements 
	}

	\author{Carlo Tiseo, Sydney Rebecca Charitos, and Michael Mistry

		\thanks{Carlo Tiseo, Sydney Rebecca Charitos and Michael Mistry are with the Edinburgh Centre for Robotics, Institute of Perception Action and Behaviour, School of Informatics, University of Edinburgh. Email: \texttt{carlo.tiseo@ed.ac.uk}}
	   \thanks{This work has been supported by the following grants: EPSRC UK RAI Hubs ORCA (EP/R026173/1), NCNR (EPR02572X/1) ; and EU Horizon 2020 project  THING (ICT-2017-1).}
	}
	\thispagestyle{empty}
\fbox{
\parbox{\textwidth}{
© 2021 IEEE.  Personal use of this material is permitted.  Permission from IEEE must be obtained for all other uses, in any current or future media, including reprinting/republishing this material for advertising or promotional purposes, creating new collective works, for resale or redistribution to servers or lists, or reuse of any copyrighted component of this work in other works.}}
\newpage
\maketitle
\begin{abstract}
The mechanism behind the generation of human movements is of great interest in many fields (e.g. robotics and neuroscience) to improve therapies and technologies. Optimal Feedback Control (OFC) and Passive Motion Paradigm (PMP) are currently two leading theories capable of effectively producing human-like motions, but they require solving nonlinear inverse problems to find a solution. The main benefit of using PMP is the possibility of generating path-independent movements consistent with the stereotypical behaviour observed in humans, while the equivalent OFC formulation is path-dependent. Our results demonstrate how the path-independent behaviour observed for the wrist pointing task can be explained by spherical projections of the planar tasks. The combination of the projections with the fractal impedance controller eliminates the nonlinear inverse problem, which reduces the computational cost compared to previous methodologies. The motion exploits a recently proposed PMP architecture that replaces the nonlinear inverse optimisation with a nonlinear anisotropic stiffness impedance profile generated by the Fractal Impedance Controller, reducing the computational cost and not requiring a task-dependent optimisation.    
\end{abstract}

\IEEEpeerreviewmaketitle

\section{Introduction}
Animals can robustly control redundant nonlinear mechanics in varying environmental conditions. In contrast, robots' interaction stability is much more fragile and less dexterous. Roboticist and neuroscientist have developed multiple theories over the years on how animals generate their motor strategies in an effort to create more interactive technology and improve the medical treatments available for people suffering from motor impairments. Recently, the increasing interest in adopting robots in a wider variety of scenarios for industrial and civil applications has further increased the interest in developing flexible and efficient control architectures. 

The two leading theory that can capture human movements are the Optimal Feedback Control (OFC), and the Passive Motion Paradigm (PMP) \cite{tommasino2017,tommasino2017task,Todorov2000,Todorov2002,guigon2007computational,shadmehr2008computational}. The OFC is probably the best-known theory capable of generating reaching trajectories and addressing the motor variability. It relies on the uncontrolled manifold approach and relies on the optimised controller to handle the local trade-off during motion execution \cite{tommasino2017,scholz1999uncontrolled}. A major issue of the OFC is that it needs a priory knowledge of the boundary conditions (i.e. equilibrium states) to formulate the optimisation problem in the presence of kinematic redundancy\cite{tommasino2017}. The Separation Principle (SP) was proposed by \textit{Guigon et al. in}\cite{guigon2007computational} to generate a human-like reaching trajectory in a redundant manipulator. SP theorises that there are two different controllers in the nervous system that work simultaneously. The first controller deals with posture dependent interaction, and optimisation while the second controller handles velocity dependant dynamic interactions. However, the theory introduces non-holonomic constraints which contrast with the experimental observation of the motor synergies \cite{guigon2007computational}. The latter are stereotyped movement strategies and were initially observed by Donders while studying eye-glazing movements, and subsequently found in other tasks (e.g., wrist pointing task and locomotion) \cite{tommasino2017,tiseo2018thesis}. The PMP is a generalisation of the Equilibrium Point Hypothesis (EPH), which relies on potential fields to model goals and mechanical constraints \cite{ivaldi1988kinematic,mohan2011passive}. Tommasino and Campolo have recently proposed the $\lambda_0$-PMP that captures the motor synergies. Their approach relies on nonlinear inverse optimisation to generate a compensatory component that introduces holonomic constraints in the joint space and generates path independent, energetic mapping of the postures \cite{tommasino2017}. The $\lambda_0$-PMP has been validated by reproducing a human-like wrist pointing task. In synthesis, the two models can generate human-like movements in redundant kinematics, but it is computationally expensive due to the nonlinear optimisations. 

Recently, we have proposed an extension to the PMP called the Harmonic-PMP (H-PMP) \cite{tiseo2021}. It is capable of generating human-like reaching trajectories. The results show how the minimum jerk trajectory observed in the task-space are consistent with harmonic trajectories distorted by the projection through the arm kinematic structure. Differently from the $\lambda_0$-PMP, the H-PMP does not require a nonlinear optimisation of the impedance to generate human-like moments. It utilises a stack of Fractal Impedance Controllers to generate a globally stable force field around the desired states. The highest controller in the hierarchy generates a smooth reaching trajectory based on the desired target position. The second uses the expected environmental forces or velocities to generate a postural optimisation of the motion. The third layer integrates this information about the interaction forces at the end-effector for tuning the joint controllers. The lowest level of the control architecture is the joint controller that receives as input: position, torque, and accuracy commands.

\begin{figure}[!htbp]
\centering
\vspace*{0.25cm}
\includegraphics[width=\columnwidth]{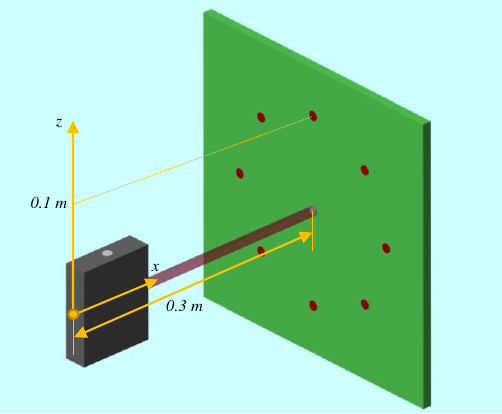}%, trim=0cm 1.5cm 0cm 2cm, clip
\caption{Simulator used to validate the proposed control architecture for the wrist point task.}
\label{fig:SimSetup}
\end{figure} 

\begin{figure}[!htbp]
\centering
\includegraphics[width=\columnwidth, trim=5cm 9.5cm 6cm 9.5cm, clip]{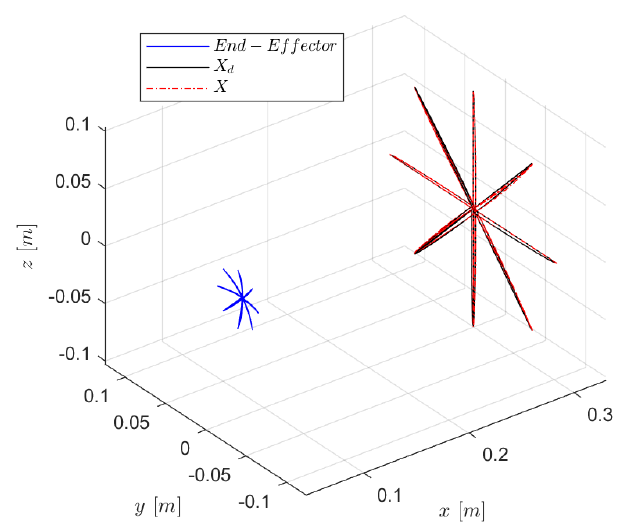}%
\caption{Example of how tracking the desired trajectory ($X_\text{d}$) projects on a spherical surface at the end-effector. The generated pointed trajectory $X$ is also shown for completeness.}
\label{fig:Example}
\end{figure}

This work evaluates the possibility that the synergies described by the Donders' law are a consequence of the spherical projection, and that a non-linear optimisation problem does not need to be solved. To prove this we will use the two layers of the H-PMP to generate torque commands to control a spherical joint in simulation (\autoref{fig:SimSetup}), which will be used the proposed spherical representation of the task (\autoref{fig:Example}). The validation of the proposed method will prove different wrist movement strategies can be generated without regressing from human data. Thus, providing insights on how humans can efficiently learn and generalise motor skills via the modulation of stable attractors defined upon deterministic nonlinear kinematics maps of the task. 

\section{Method}
A physics simulation of the wrist joint is developed using the Simscape Simulink Library (Mathworks Inc, US), which uses a \texttt{ode45} solver. The simulations are run on an Intel i7-7700HQ processor with \SI{16}{\giga\byte} of RAM. The quaternion representation is used for projecting the target on the spherical surface, which has already been used for recording measuring the eye movements \cite{tweed1990computing}. 

\subsection{Fractal Impedance Controller}
The fractal impedance controller is an anisotropic conservative attractor that ensures upper-bounded forces, velocities and power \cite{tiseo2020bio,tiseo2021}. The system dynamics is determined starting from a continuous finite effort (force/torques) profile ($F_\text{d}$) acting on the system inertia.
\begin{equation}
F=\left\{\begin{array}{cc}
     F_\text{d} \left( \tilde{x}\right) &  \text{divergence}\\
     \frac{2F_\text{d}\left(\tilde{x}_\text{max}\right)}{\tilde{x}_\text{max}} \left( \tilde{x}-\frac{\tilde{x}_\text{max}}{2} \right)&  \text{convergence}
\end{array} \right.
    \label{FICBasic}
\end{equation}
where $\tilde{x}$ is the end-effector distance from the desired pose and $\tilde{x}_\text{max}$ is the maximum distance reached during the divergent phase. 

\subsection{Quaternions and Listing's Planes}
Listing's planes are the non-torsional motion planes identified in the eye motor synergies. The quaternion representation of this motion are usually obtained via the rotation matrices \cite{tweed1990computing,campolo2011pointing}. We propose a representation based on projecting the planar task on a spherical surface. Simultaneously, the torsional degree of freedom is controlled to align the tangent-space with the task.  Thus, the absence of torsion during movements implies a constant orientation ($\phi$) around $\hat{r}$ (i.e., absence of roll), as shown in \autoref{fig:Example}. The vector $\hat{r}$ can be derived using the positions of the base frame of the spherical joint ($x_0$) and the end-effector/target Cartesian coordinates ($x$). 
\begin{equation}
   \left.\begin{array}{l}
        \hat{r}=\cfrac{\vec{x}-\vec{x}_\text{0}}{||\vec{x}-\vec{x}_\text{0}||}  \\
        c_\theta=\left[1~0~0\right] \cdot \hat{r} \\
        \text{if}~c_\theta =-1 \\
        ~~~q_0=\left(0~\hat{n}\right)\\
        \text{else} \\
        ~~~q_0=\left(\left(1+c_\theta\right)~ \left[1~0~0\right]^\text{T}\times \hat{r} \right)\\
         \text{end}\\
         q=\cfrac{\left(\cos(\phi/2)~\left[\sin(\phi/2)~0~0\right]^\text{T}\right)q_0}{||q_0||}\\
   \end{array}\right.
    \label{quaternion}
\end{equation}
where $\hat{n}$ is a vector lying the plane orthogonal to $\hat{r}$.  

\subsection{Harmonic Passive Motor Paradigm}
\begin{figure}[!htbp]
\centering
\includegraphics[width=\columnwidth, trim=0.5cm 1.5cm 0.5cm 1.5cm, clip]{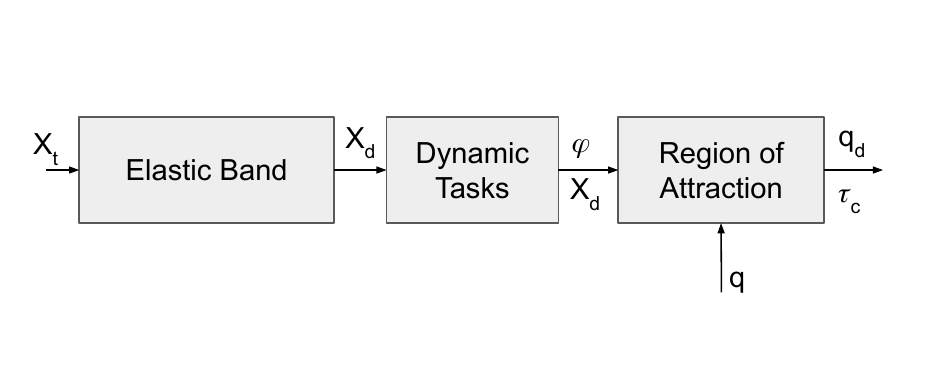}
\caption{The elastic band generates a smooth trajectory between the initial position and the desired target ($X_\text{t}$). The dynamic task exploits the redundancy to align the internal dynamics with the expected environmental dynamics selecting $\phi$. The information of the environmental interaction is also used to set a reference tuning for the region of attraction. The dynamics task also selects the initial value for the controller stiffness ($K$), which can be adjusted based on the sensorial feedback. The region of attraction uses the desired position ($X_\text{d}$), the current orientation ($q$), and the selected $\phi$ and $K$ to generate a torques command ($\tau_\text{d}$) and the desired orientation ($q_\text{d}$).}
\label{fig:CtlArc}
\end{figure}

The H-PMP architecture extends from the tasks-space planning to the low-level interaction controllers at the joints. However, this work utilises only the Elastic Band, Dynamic Task and Region of Attraction modules \autoref{fig:CtlArc} \cite{tiseo2021}. The choice has been made to render the presented results independent from the kinematics used to implement the spherical joint. Thus, the end-effector torques can be directly used as a control input to the spherical joint being the degrees of freedom aligned with the end-effector frame \cite{siciliano2010robotics}. 

The desired trajectory is obtained from the distance between the end-effector and the target position at the beginning of the movement \cite{tiseo2020Planner,tiseo2021}.

\begin{equation}
X_d=\iint_{t_0}^{t} \ddot{X}_\text{d} dt^2=\iint_{t_0}^{t} \frac{K_\text{d}\left(X_\text{d}\left(t-1\right)-X_\text{t}\right)}{M_\text{d}} dt^2
    \label{EqElasticBand}
\end{equation}

where $X_\text{t}$ is the current target location. $K_\text{d}$ is the stiffness of the elastic band determined based on the desired maximum acceleration \cite{tiseo2021}. $M_\text{d}$ is the desired inertia for the end-effector. The Cartesian trajectory obtained by \autoref{EqElasticBand} is transformed in quaternion using \autoref{quaternion}.

A FIC controller is used to drive the wrist from the current pose to the desired pose. The intrinsic stability of the FIC enables to define of the field without worrying about the stability \cite{babarahmati2019,tiseo2020}, which is particularly useful when using a redundant representation for space (i.e., quaternion). In contrast, when relying on a traditional formulation of the impedance controller, the dynamic parameters have to be re-optimised to be asymptotically stable every-time the controller dynamics is adjusted \cite{Angelini2019}.  

The FIC controller in quaternion representation can be defined starting from a single stiffness parameter, and the quaternion of the current and desired end-effector poses. The control output is the torque in the base frame, which can be used directly to control the ideal spherical joint used in this work. 

\begin{equation}
\begin{array}{l}
     \tilde{q}=q_\text{d}q^{-1}\\
     \theta=2\arctan(||\tilde{q}_\text{v}||,\tilde{q}_\text{s})\\
     \tau_\text{cq}=\left\{\begin{array}{cc}
          K\theta&  ~\text{divergence}\\
          2K\left(\theta-\cfrac{\theta_\text{max}}{2}\right)&  ~\text{convergence}
     \end{array}\right.\\
     \tau_\text{c}=\mathrm{sign}\left(\tilde{q}_\text{s}\right)\tau_\text{cq}\tilde{q}_\text{v}
\end{array}
\label{FICquaternion}
\end{equation}
where $\tilde{q}_\text{v}$ is the vector component of the quaternion, $\tilde{q}_\text{s}$ is the scalar component of the quaternion, K is a constant stiffness. It is worth noting that these control torqued will require to be projected in the joint space of the chosen mechanical when deployed on a real structure, using the Jacobian transposed \cite{siciliano2010robotics}.

\subsection{Experimental set-up}
The wrist pointing task used to evaluate the $\lambda_0$-PMP is a planar clock experiment on the \textit{yz}-plane which has eight targets \cite{tommasino2017,tommasino2017task}. The clock design is adapted from an exercise commonly used in robotic rehabilitation therapies and collected using a high-transparency \cite{campolo2010kinematic,campolo2011pointing}. The method was used both with multiple devices with different levels of transparencies \cite{tagliamonte2011effects,campolo2009intrinsic,campolo2010kinematic,campolo2011pointing}. The comparative data reported in \cite{tagliamonte2011effects} indicate that the subjects' pointing strategies tend to converge with the decrees of the measuring set-up transparency. 

The targets plane is at a distance of \SI{.30}{\meter} from the origin along the \textit{x}-axis. They are equally spaced on a circumference having a radius of \SI{0.10}{\meter} and centred in $P=(0.3, 0, 0)~\si{\meter}$.  All the collected data are sampled at \SI{1}{\kilo\hertz}. The wrist is modelled as a spherical joint to show that the result is independent of the joint sequence of the mechanism. 

Experiments are conducted with and without considering the gravitational field to verify that the FIC can compensate the nonlinear dynamics even using the quaternion representation. The hand mass \SI{1}{\kilo\gram} and the inertia matrix is calculated based on the geometry of the rectangle used in the simulation. The dimensions of the hand are H= \SI{.1}{\meter}, L= \SI{.08}{\meter} and T= \SI{.02}{\meter}. The maximum acceleration used to compute $K_\text{d}$ is \SI{3.2}{\meter\per\second^2}. The user manually adjusts the value of $K$ and $\phi$. The Listing's plane is generated assuming that the joints are disposed on a configuration consistent with $\mathrm{XYZ}$ euler angles representation. This convention is chosen based on the arrangement of the degrees of freedom in the 
human wrist \cite{campolo2009intrinsic}.

\section{Results}
\begin{figure}[!htbp]
\centering
\includegraphics[width=\columnwidth, trim=5cm 9.25cm 5.5cm 8.75cm, clip]{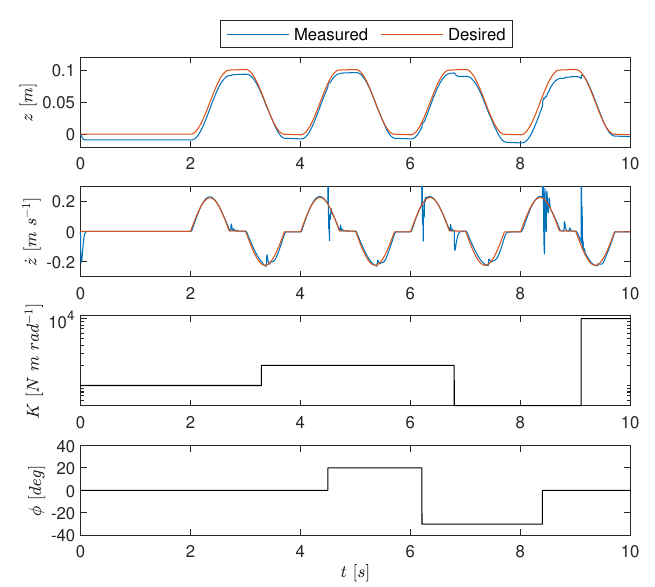}%
\caption{The end-effector trajectory is smooth, the velocities are bell-shaped, and both $K$ and $\phi$ can be changed online without affecting the system stability.}
\label{fig:MultTraj}
\end{figure}

The first experiment conducted is to verify if the planned motion are smooth and if $\phi$ and $K$ can be adjusted online. The initial value selected for $K$ is \SI{1}{\kilo\newton\meter\per\radian} and the task was limited to the top target. The data showed in \autoref{fig:MultTraj} show that both the planned and the end-effector trajectory are smooth, they have the typical bell-shaped velocity profiles, and both $\phi$ and $K$ can be changed online without destabilising the system. 

\begin{figure}[!htbp]
\centering
\includegraphics[width=\columnwidth, trim=5cm 9.5cm 5.5cm 9.5cm, clip]{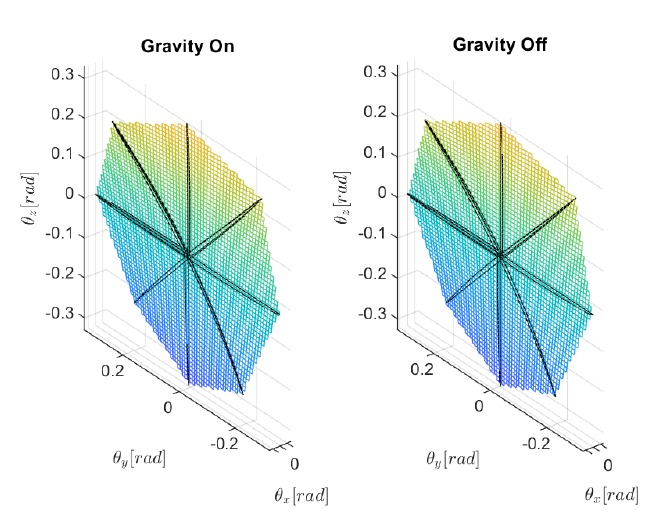}%
\caption{The comparison of the Listing's plane obtained with and without gravity with a rigid controller shows how the controller can compensate for the deformation generated by gravity.  Our surfaces a flatter compared to the human data reported in \cite{campolo2011pointing}, because the joint mechanics is ideal in our simulations.}
\label{fig:NoGListing}
\end{figure}

The Listing's plane for measured with and without gravity (\autoref{fig:NoGListing}) are obtained setting $K=$\SI{10}{\kilo\newton\meter\per\radian} and a $\phi=0$ \si{\radian}. The average effort during the task increases from $0.183 \pm 0.158$\si{\newton\meter} to $0.423 \pm 0.165$\si{\newton\meter}. The Root Mean Square Error (RMSE) along $z$ increases from $2$ to $3$ \si{\milli\meter}, while it remains stable to \SI{1}{\milli\meter} along $y$.

\begin{figure}[!htbp]
\centering
\includegraphics[width=\columnwidth, trim=5cm 9.5cm 5.5cm 9cm, clip]{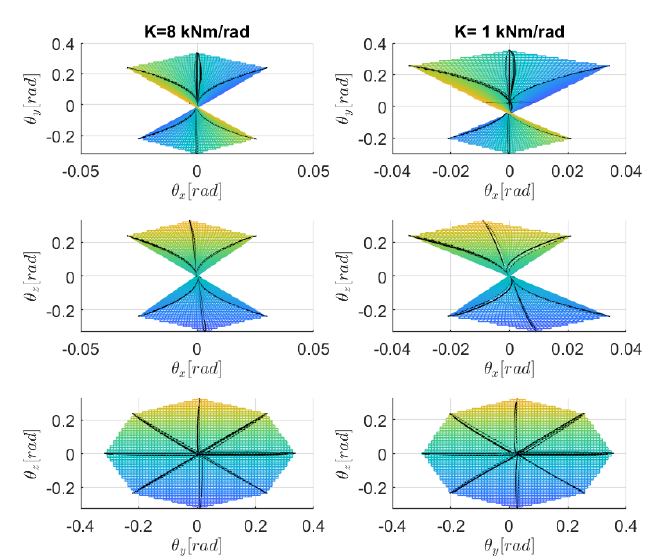}%
\caption{Reducing the controller rigidity introduces a deformation in the measured Listings' plane due to the gravitational pull that is not representative of a change in strategy of the controller.}
\label{fig:G18KListing}
\end{figure}

We repeated the experiment with gravity reducing the stiffness to $K=$\SI{8}{\kilo\newton\meter\per\radian} and $K=$\SI{1}{\kilo\newton\meter\per\radian}. Under this conditions there is a negligible change in the average effort made $0.425 \pm 0.165$\si{\newton\meter} for $K=$\SI{1}{\kilo\newton\meter\per\radian}, and $0.426 \pm 0.165$\si{\newton\meter} for $K=$\SI{8}{\kilo\newton\meter\per\radian}. In contrast, the tracking error increases becoming RMSE$_y=$\SI{4}{\milli\meter} and RMSE$_z=$\SI{9}{\milli\meter} for the softer controller, and RMSE$_y=$\SI{2}{\milli\meter} and RMSE$_z=$\SI{4}{\milli\meter} for the more rigid controller. The data also show a deformation of the measured Listing's plane affecting the planes where is involved $\theta_x$ (\autoref{fig:G18KListing}), despite there is not a change in the planned strategy. Thus, it indicates that it is connected to the reduced effectiveness in compensating the external force field. 

\begin{figure}[!htbp]
\centering
\includegraphics[width=\columnwidth, trim=5cm 9.5cm 5.5cm 9cm, clip]{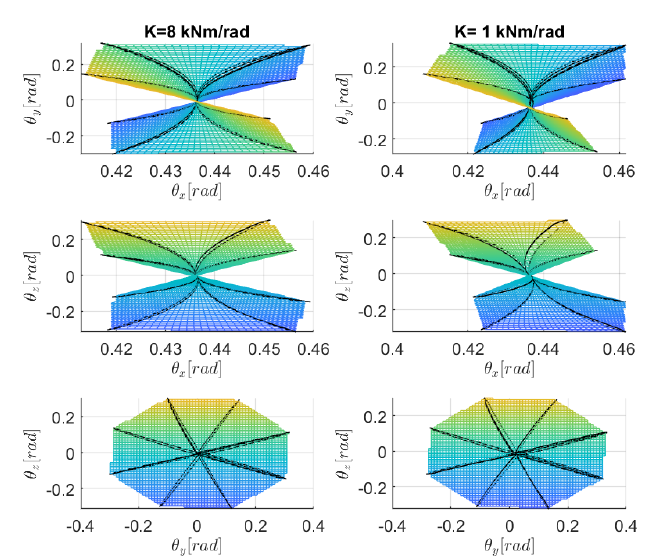}%
\caption{Comparing the Listing's planes generated for $\phi=\SI{-25}{\deg}$, we can observe a visible compared to the shape of the planes in \autoref{fig:G18KListing}. However, they also share a similar deformation patter when reducing the stiffness is reduced without changing the desired behaviour.}
\label{fig:PhiN25G18KListing}
\end{figure}

Changing $\phi$ to \SI{-25}{\deg} in the scenario described in the previous paragraph, revealed that the RMSE remained the same for both the values of the stiffness $K$. RMSE$_y=$\SI{4}{\milli\meter} and RMSE$_z=$\SI{9}{\milli\meter} for the softer controller, and RMSE$_y=$\SI{2}{\milli\meter} and RMSE$_z=$\SI{4}{\milli\meter} for the more rigid controller. Similarly, the average effort is within the standard deviation in both cases. The values are $0.430 \pm 0.166$\si{\newton\meter} for $K=$\SI{1}{\kilo\newton\meter\per\radian} and to $0.427 \pm 1.66$\si{\newton\meter} for $K=$\SI{8}{\kilo\newton\meter\per\radian}. Although \autoref{fig:PhiN25G18KListing} describes a different Listing's plane than \autoref{fig:G18KListing}, a similar deformation patter can be observed with the reduction of the  due to the change of the controller stiffness

Comparing \autoref{fig:G18KListing} and \autoref{fig:PhiN25G18KListing} shows that change in $\phi$ induced a rotation of the Listing's plane around $\theta_{x}$ and it requires a changed in the planned strategy while changing the tuning of the controller and changing its efficacy in counteracting an external force deforms the plane in the direction $\theta_y$ and $\theta_z$.

\section{Discussion}
The experiments show the proposed method can accurately control the wrist pointing task without solving the nonlinear inverse optimisation. The show that the Listing's planes deformation is both caused by changes in the spherical projection (i.e., change in the posture) and a consequence of the intrinsic impedance of the wrist as previously hypothesised \cite{campolo2009intrinsic,campolo2010kinematic,campolo2011pointing,tommasino2017,tommasino2017task}. Therefore, it is impossible to decouple the two phenomena when performing a nonlinear inverse optimisation on the measured human movements, rendering their solution task-specific and not possible to be generalised. In contrast, the proposed method allows decoupling the two problems, allowing to independently address the dynamic interaction and the postural optimisation by tuning $K$ and selecting $\phi$, respectively. It can also be argued that the absence of torsion in the wrist in the pointing experiment is task-specific, and there are other activities where the torsional components are essential—for example, introducing spin when throwing a ball or playing tennis. These actions require changing Listing's plane ($\phi\neq\mathrm{constant}$) during the motion to spin the ball. The modelling of these activities above using the $\lambda_0$-PMP would require the continuous optimisation of the impedance to model due to the changing task-space projector.

The observations made in the previous paragraph suggest that the torsion ($\phi$) is used to align the anisotropic properties of the body with the task to optimise the distribution of the effort in the joints by taking advantage of the kineto-static duality, which is a principle commonly used in robotics \cite{siciliano2010robotics}. Based upon this interpretation, the selection of a Listing's plane is equivalent to optimise the static control of the SP proposed in \cite{guigon2007computational}. Thus, the motor synergies can be interpreted as a policy that exploits the redundant degree of freedom policy to align the tangent space with the task without requiring solving the inverse problem, as it is also shown for the 3-link arm in \cite{tiseo2021}. However, this method cannot be currently be extended to a 7-DoF arm due to the lack of an alternative solution to inverse kinematics optimisations proposed in \cite{tiseo2020,tiseo2020bio}. 

The dynamic controller of the Separation Principle arises the mediation of the FIC controller in the Region of Attraction \autoref{fig:CtlArc} \cite{tiseo2021}, exploiting the intrinsic characteristics of the FIC to synchronise with the external dynamics. Nevertheless, this would require a careful selection of $X_t$ to accounts for the energetic manifold of the environment, as hypothesised in \cite{tiseo2018thesis} based on the observation that human locomotion is synchronised with the gravitational manifold \cite{tiseo2018bipedal}. These observations are also confirmed by the finding of more recent experiments, where researchers proved how the performances during dynamic tasks are correlated with the synchronisation with the attractors of different tasks \cite{Zhang2018,Sohn2020}. It is worth noting that the elastic band has a similar function to the cost-to-go model in \cite{shadmehr2008computational}. The FIC controller of the elastic band generates a potential field per unit of mass in agreement with the EPH. However, this also implies that the proposed method will have similar limitations to the one described for the minimum jerk planning in \cite{shadmehr2008computational}, if the $X_t$ and $\phi$ are selected without accounting for the environmental dynamics.

The deformation of Listing's plane for different tuning of the controller when interacting with gravity is comparable to the shift observed in human subjects when interacting with different devices \cite{tagliamonte2011effects}. The data \autoref{fig:NoGListing} show that when the controller is sufficiently rigid, the behaviour is equivalent to the one obtained without gravity. The reduction of the controller rigidity allows the gravitational pull to deform the planned synergy described by $q_d$ in the synergy measured at the end-effector ($q$). These results provide an alternative explanation to the findings described by \cite{tagliamonte2011effects} that can be now explained a change in the interaction rather than a changed in planned strategy. We suggest the interpretation of $K$ as the trade-off between tracking accuracy and effort of the UMH, which can now be performed without nonlinear inverse optimisation \cite{scholz1999uncontrolled,guigon2007computational,tommasino2017, Todorov2000}.  

We would also like to propose an interpretation of the FIC in terms of neural circuits, which still requires further validation. The FIC defines an adaptable conservative energy field that ensures smooth convergence to the desired state along its harmonic autonomous trajectories \cite{babarahmati2019,tiseo2020, tiseo2021}. The attractor can be described as a van der Pol Oscillator as shown in Appendix, which can be implemented using Central Pattern Generator (CPG) neural networks \cite{Ohgane2009,Bucher2015}. CPGs are traditionally used to model oscillatory circuit controlling rhythmic behaviour in the spinal cord and brain stem, but there is also evidence that "learning" CPGs are present in the motor cortex, and they have higher plasticity \cite{yuste2005cortex}. Therefore, it is possible that adaptable cortical CPGs could implement a behaviour similar to the FIC in the proposed architecture.

\section{Conclusion}
The proposed architecture can integrate multiple aspects of the principal theories on human motor control into a single architecture. It is the only method that does so without requiring a nonlinear inverse optimisation to identify the controllers' parameters. Furthermore, the stability properties of the FIC also enable online tuning of the controller stiffness $K$, enabling to adjust the tracking performances acting on a single parameter to adapt to change in the environmental interaction. Concurrently, the selection of $X_t$ and $\phi$ allow preparing the motor strategy to interact with expected interactions with the environment. Future work will focus on extending these results to 7-DoF manipulators and legged locomotion.

\appendix
\label{appendix}
The FIC is a conservative controller generating an attractor with harmonic autonomous trajectory, and it can also be described with a Li\`enard equation that describes systems having a nonlinear friction term. The van der Pol oscillator (VPO) is probably the known example of a Li\`enard equation \cite{strogatz2018nonlinear}. Thus, we will use that representation to show that the FIC generates a dynamics consistent with the Li\`enard equations ($\ddot{x}+f(x)\dot{x}+g(x)=0$) \cite{strogatz2018nonlinear}. The VPO equation is:
\begin{equation}
  M\ddot{x}=\mu \left(1-x^2\right)\dot{x}-Kx
\label{VDP}
\end{equation}
where $M$ is the mass, $\mu$ is a friction coefficient, and $K$ is the stiffness. 

The autonomous trajectory of the FIC are designed imposing the conservation of energy at the switching between divergence and convergence \cite{babarahmati2019}, and this condition can be used to identify the $\mu$ value for the VPO associated with every trajectory of the attractor. The first step is to derive the relationship between maximum displacement ($\tilde{x}_\text{max}$) and maximum velocity ($\dot{x}_\text{max}$), using the conservation of energy.
\begin{equation}
\begin{array}{l}
   \Delta U_\text{max}=\Delta E_\text{max}\\\\
   \Delta U_\text{max}=\cfrac{1}{2}K\left(\tilde{x}_\text{max}\right)\tilde{x}_\text{max}^2\\\\
   \Delta E_\text{max}=M\dot{x}_\text{max}^2\\\\
\dot{x}_\text{max}=\sqrt{\cfrac{K\left(\tilde{x}_\text{max}\right)}{2M}}\tilde{x}_\text{max}^2=\omega_n (\tilde{x}_\text{max})\tilde{x}_\text{max}
\end{array}
\label{FIC-A}
\end{equation}
The $\mu$ value of the equivalent VPO can be imposing the conservation of energy and the convergence to the desired state and substituting the relationship between $\tilde{x}_\text{max}$ and $\dot{x}_\text{max}$ obtained in \autoref{FIC-A}. 
\begin{equation}
\begin{array}{l}
       M\dot{x}_\text{max}^2 + \cfrac{1}{2}K\left(\tilde{x}_\text{max}\right)\tilde{x}_\text{max}^2 +\mu \int_0^{\tilde{x}_\text{max}} (1-\tilde{x}^2)\dot{\tilde{x}} d\tilde{x}\\\\
       \mu(\tilde{x}_\text{max})= \cfrac{M\omega_n^2 (\tilde{x}_\text{max})\tilde{x}_\text{max}^2 + K\left(\tilde{x}_\text{max}\right)\tilde{x}_\text{max}^2}{2\int_0^{\tilde{x}_\text{max}} (1-\tilde{x}^2)\dot{\tilde{x}} d\tilde{x}}
\end{array}
\label{VPO-A}
\end{equation}
However, this leads to a jump in force when using a nonlinear stiffness, which has been solved imposing the conservation of force as switching condition. The jump in energy ($E_\text{VA}$) introduced by imposing the force conservation is accounted by a virtual antagonist \cite{tiseo2020bio}. Thus, $E_\text{VA}$ needs to be added to the energy conservation equation in \autoref{VPO-A}.
\begin{equation}
      \mu(\tilde{x}_\text{max})= \cfrac{M\omega_n^2 (\tilde{x}_\text{max})\tilde{x}_\text{max}^2 + K\left(\tilde{x}_\text{max}\right)\tilde{x}_\text{max}^2+E_\text{VA}}{2\int_0^{\tilde{x}_\text{max}} (1-\tilde{x}^2)\dot{\tilde{x}}d\tilde{x}}
\label{VPO-B}
\end{equation}
Therefore, the FIC is just an efficient algorithmic representation of an asymptotically stable Li\`enard system. 
\balance
\bibliography{main}
\bibliographystyle{IEEEtran}
\end{document}